\documentclass{article}

\usepackage{arxiv}

\usepackage[utf8]{inputenc} 
\usepackage[T1]{fontenc}    
\usepackage{hyperref}       
\usepackage{url}            
\usepackage{booktabs}       
\usepackage{amsfonts}       
\usepackage{nicefrac}       
\usepackage{microtype}      
\usepackage{lipsum}
\usepackage{graphicx}
\usepackage{natbib}
\usepackage{amsmath}
\usepackage{amssymb}

\usepackage{multirow}
\usepackage{amsthm}

\title{ReSAE: Residualized Sparse Autoencoders for Multi-Layer Transformer Interventions}

\author{
 Prathyush Poduval$^{*,a}$, Calvin Yeung$^{a}$, Neel Desai$^{b}$, Mohsen Imani$^a$ \\
  $^a$Department of Computer Science\\
  University of California, Irvine\\
  Irvine, CA 92697 \\
  $^b$Georgia Institute of Technology\\
  Atlanta, GA 30332\\
  $^*$Corresponding author: \texttt{ppoduval@uci.edu} \\
}

\begin{document}
\maketitle
\begin{abstract}
Sparse autoencoders are usually trained one layer at a time, even though transformer residual
stream activations are strongly coupled across depth. This creates a practical problem for
multi-layer interventions: different layerwise dictionaries can spend capacity representing the
same carried-forward information, and replacing several layers at once can produce interactions
that are not predicted by single-layer behavior. We introduce Residualized Sparse Autoencoders
(ReSAEs), which fit an affine map between selected layers and train each later-layer SAE on the
unexplained residual rather than on the full activation. Reconstructions are mapped back into the
original activation space through the fitted affine chain, so ReSAEs can be evaluated with the
same intervention protocols as ordinary SAEs. On Pythia-1.4B and Gemma-2-9B, residualization reduces decoder redundancy and improves sparse probing and targeted perturbation in most tested settings. Despite reconstructing less of the raw activation variance, ReSAEs recover more transformer cross entropy under multi-layer replacement. This gain is clearest under teacher-forcing and at sufficient sparsity online, indicating that ReSAEs preserve the components of the activation most relevant to the model's downstream computation. These results suggest that removing linearly predictable cross-layer structure is a useful default for multi-layer SAE interventions.
\end{abstract}

\section{Introduction}
\label{sec:intro}

As large language models (LLMs) scale in capability and deployment, understanding their internal representation has become a major point of research. A range of poorly understood behaviors like hallucinations~\cite{ji2023survey}, jailbreaks~\cite{zou2023universal}, emergent misalignment~\cite{betley2025emergent}, and sycophancy~\cite{sharma2023sycophancy} remain difficult to predict or prevent without mechanistic understanding of how information is structured inside these models. The goal of interpretability is to identify what representations the model builds, how they evolve across layers, and how interventions on them can be made reliable.

\begin{figure*}[t]
  \centering
  \includegraphics[width=\textwidth]{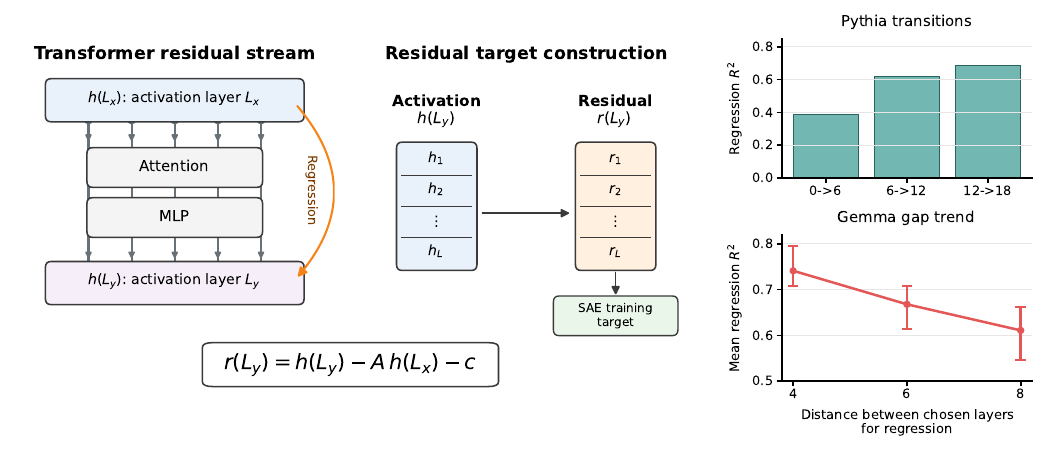}
  \caption{Residualization overview and cross-layer regression $R^2$. \textbf{Left}: selected
           transformer activations are connected through the residual stream, with an affine
           regression mapping an earlier activation $h(L_x)$ to a later activation $h(L_y)$.
           \textbf{Middle}: ReSAEs train the later-layer SAE on the residual
           $r(L_y)=h(L_y)-A h(L_x)-c$ rather than the full activation. \textbf{Right}: affine
           regressions explain a large fraction of downstream activation variance in Pythia-1.4B
           transition pairs and in Gemma-2-9B, where mean $R^2$ decreases as the distance between
           chosen layers increases.}
  \label{fig:layer_correlations}
\end{figure*}

Sparse autoencoders (SAEs) have emerged as a central tool for this goal. An SAE decomposes a transformer's internal activations into sparse combinations of directions drawn from a learned dictionary. These individual active directions have been shown to correspond to interpretable, monosemantic concepts in many cases~\cite{bricken2023monosemanticity,cunningham2023sparse}. Several variants improve upon the basic formulation: BatchTopK~\cite{gao2025scaling} and JumpReLU~\cite{rajamanoharan2024jumping} improve sparsity control and training stability, p-Anneal~\cite{karvonen2024measuring} achieves better decomposition at lower sparsity, and Matryoshka SAEs~\cite{bussmann2025matryoshka} learn monosemantic and hierarchical features. Large-scale public releases like GemmaScope~\cite{lieberum2024gemmascope}, LlamaScope~\cite{wu2024llamascope}, and SAELens~\cite{bloom2024saelens} have trained SAEs across many layers of frontier-scale models, establishing sparse feature dictionaries as a practical vocabulary for inspecting model internals.

Training SAEs independently at each layer, however, ignores a structural property of the transformer residual stream: each layer is not independent of the ones before it. Each sub-module of the architecture can be thought of as reading from some subspace of the residual stream and writing an update back into it~\cite{elhage2021mathematical}, so a substantial fraction of any given layer's activation is information carried forward or lightly transformed from earlier layers~\cite{raghu2017svcca, kornblith2019similarity}. Figure~\ref{fig:layer_correlations} demonstrates that simple affine regressions between selected layers explain a large fraction of downstream activation variance in both Pythia-1.4B and Gemma-2-9B. Therefore, training per-layer SAEs independently on raw activations wastes capacity relearning this predictable carried-forward structure. Moreover, multi-layer interventions become difficult to implement, since if the same information appears in multiple layerwise dictionaries, simultaneous feature replacement can produce non-additive effects as features duplicate, cancel, or drive activations out of the SAE's training distribution across layers.

We propose \textit{Residualized Sparse Autoencoders} (ReSAEs) to address this. For each consecutive pair of selected layers, we fit an affine map from the earlier layer's activation to the later one, and train the later-layer SAE on the residual, which is the component not explained by this map. At reconstruction time, the SAE output is composed with the fitted affine chain to recover the original activation space, keeping ReSAEs directly comparable to standard SAEs at evaluation. This forces each SAE to encode what is new at its layer rather than what is already predictable from the layer below. Moreover, the effect of linear interventions performed at one layer can directly be viewed at the next layer through this affine transformation. 

Our experiments show that residualization improves multi-layer SAE behavior in the intended direction. ReSAEs reconstruct a strictly harder target (reflected in lower raw explained variance) yet recover transformer cross entropy more faithfully under simultaneous multi-layer intervention, produce less redundant decoder directions, and outperform raw SAEs on downstream sparse probing and targeted perturbation {evaluations}. Our contributions are:

\begin{enumerate}
  \item We introduce Residualized Sparse Autoencoders (ReSAEs), which train each per-layer SAE on the component of its layer's activation not linearly predictable from the previous selected layer, and compose SAE reconstructions with a fitted affine chain to recover original activation space.
  \item We show that ReSAEs recover more cross entropy under multi-layer intervention despite lower explained variance, and produce more orthogonal decoder directions (evidence that removing predictable cross-layer structure directs SAE capacity toward layer-local information that is more useful for intervention).
  \item We report evaluations across reconstruction quality, cross-entropy degradation, decoder similarity, sparse probing, targeted probe perturbation, and spurious correlation removal on Pythia-1.4B and Gemma-2-9B across multiple sparsity levels.
\end{enumerate}

\section{Prior Work}
\label{sec:prior}

\paragraph{Sparse Autoencoders for Mechanistic Interpretability.}
Sparse autoencoders (SAEs) have emerged as a central tool for decomposing transformer hidden states into interpretable feature directions~\cite{cunningham2023sparse,bricken2023monosemanticity}. Early work trains ReLU SAEs on individual layers of small language models and shows that individual latent directions correspond to interpretable token or context properties. BatchTopK~\cite{gao2025scaling} and JumpReLU~\cite{rajamanoharan2024jumping} variants improve training stability and feature utilization, and are now standard for large-scale sweeps. These approaches all train on a \textit{single} selected layer; the multi-layer setting is largely
unexplored.

\paragraph{Multi-Layer and Hierarchical SAEs.}
Matryoshka SAEs~\cite{bussmann2025matryoshka} train nested groups of latents across layers by sharing a prefix of latent codes, allowing coarse-to-fine decomposition. Cross-Coders~\cite{lindsey2024sparsecrosscoders} jointly train on pairs of adjacent layers to identify features that are shared across layers. Both approaches use raw activations and are susceptible to the cross-layer redundancy problem that
motivates this work.

\paragraph{Cross-Layer Redundancy and Interventions.}
Hidden states across transformer layers are strongly correlated~\cite{raghu2017svcca, kornblith2019similarity}. This redundancy has been studied for transfer learning and representation compression, but its impact on SAE intervention quality has received little attention. \citet{elhage2021mathematical} characterize superposition and residual stream structure in transformers, motivating a separation of ``new'' versus ``carried-forward'' information. Our work operationalizes this separation via affine regression between selected layers, a tool previously used for layer similarity analysis but not for SAE training targets.

\paragraph{Intervention Stability.}
Online SAE interventions (replacing a model's activations with SAE reconstructions during a forward pass) accumulate errors across layers because each SAE sees an out-of-distribution input caused by upstream replacement errors. \citet{templeton2024scaling} and subsequent work note that single-layer interventions degrade downstream computation. {Multi-layer simultaneous intervention is an open problem and we use a simple interaction-residual metric, explained in Section~\ref{sec:eval-metrics}, to measure whether single-layer intervention results predict multi-layer behavior.}

\section{Methodology}
\label{sec:method}

\subsection{Problem Setup}

Let $L = (\ell_1, \ell_2, \ldots, \ell_M)$ be a strictly increasing subset of transformer layer
indices, $\ell_1 < \ell_2 < \cdots < \ell_M$.
For token position $i$ in sequence $b$, let $h^{(\ell_m)}_{b,i} \in \mathbb{R}^{d}$ denote the hidden
state at selected layer $\ell_m$.
The goal is to train $M$ independent sparse autoencoders, one per selected layer, such that:
\begin{enumerate}
  \item each SAE encodes only the information \textit{newly} introduced at its layer rather than information carried forward from earlier selected layers;
  \item multi-layer online interventions compose without compounding error.
\end{enumerate}

\subsection{Pairwise Affine Residualization}

For each consecutive pair $(\ell_m, \ell_{m+1})$ in $L$, we fit an affine map
\begin{equation}
  \begin{aligned}
    h^{(\ell_{m+1})}_{b,i} &\approx A_m h^{(\ell_m)}_{b,i} + c_m,\\
    A_m &\in \mathbb{R}^{d \times d}, \qquad c_m \in \mathbb{R}^d,
  \end{aligned}
  \label{eq:affine}
\end{equation}
{using ridge-regularized ordinary least squares on a held-out calibration set; implementation details, including calibration-set size, are given in Section~\ref{sec:setup}.}
The residualized activation at layer $\ell_{m+1}$ is
\begin{equation}
  r^{(\ell_{m+1})}_{b,i}
  \;=\; h^{(\ell_{m+1})}_{b,i} - \bigl(A_m h^{(\ell_m)}_{b,i} + c_m\bigr).
  \label{eq:residual}
\end{equation}
The regression chain is defined over the chosen subset $L$, not over all model layers.
For the anchor layer $\ell_1$, we center by the empirical mean:
$\tilde{h}^{(\ell_1)} = h^{(\ell_1)} - \mu_1$, where $\mu_1 = \mathbb{E}[h^{(\ell_1)}]$.

\subsection{Block Normalization}

Residual blocks have different norms at different depths, so we apply per-block RMS normalization:
\begin{equation}
  S_m = \frac{1}{\sigma_m}, \qquad
  \sigma_m = \sqrt{\mathbb{E}\!\left[\|r^{(\ell_m)}\|_2^2 / d\right] + \varepsilon}.
  \label{eq:norm}
\end{equation}
The normalized input to the SAE at block $m$ is $z^{(m)} = S_m r^{(\ell_m)}$ (with
$z^{(1)} = S_1 \tilde{h}^{(\ell_1)}$ for the anchor).

\subsection{Independent Per-Layer SAE Training}

Each block trains its own BatchTopK SAE~\cite{gao2025scaling} with its own encoder and decoder.
For block $m$, the SAE model is
\begin{equation}
  \begin{aligned}
    a^{(m)}_{b,i}
    &= W^{(m)}_e z^{(m)}_{b,i} + b^{(m)}_e,\\
    f^{(m)}_{b,i}
    &= \operatorname{TopK}_k\!\left(\operatorname{ReLU}\!\left(a^{(m)}_{b,i}\right)\right),\\
    \hat{z}^{(m)}_{b,i}
    &= W^{(m)}_d f^{(m)}_{b,i} + b^{(m)}_d,
  \end{aligned}
  \label{eq:sae}
\end{equation}
where $W^{(m)}_e,b^{(m)}_e$ are the encoder parameters, $W^{(m)}_d,b^{(m)}_d$ are the decoder
parameters, and $\operatorname{TopK}_k$ keeps only the largest $k$ nonnegative activations. The full latent is
$f_{b,i} = (f^{(1)}_{b,i}, \ldots, f^{(M)}_{b,i})$.
The per-block loss is the reconstruction objective under this hard sparsity constraint:
\begin{equation}
  \begin{aligned}
    \mathcal{L}^{(m)}_{\text{SAE}}
    &=
    \mathbb{E}_{b,i}\!\left[
      \frac{1}{d}\left\|z^{(m)}_{b,i} - \hat{z}^{(m)}_{b,i}\right\|_2^2
    \right],\\
    \left\|f^{(m)}_{b,i}\right\|_0 &\le k .
  \end{aligned}
  \label{eq:sae_loss_block}
\end{equation}
The total objective is blockwise separable:
\begin{equation}
  \mathcal{L}_{\text{SAE}}
  = \sum_{m=1}^{M} \mathcal{L}^{(m)}_{\text{SAE}},
  \label{eq:loss}
\end{equation}
where gradients from block $m'$ do not update the SAE for block $m \neq m'$.
All selected layers share a single activation-collection pass for computational efficiency, but
the SAE objectives and parameters remain fully layer-local.

\subsection{Original-Space Reconstruction}

After encoding, we recursively reconstruct original-space activations:
\begin{align}
  \hat{h}^{(\ell_1)}_{b,i}
  &= \mu_1 + S_1^{-1}\hat{z}^{(1)}_{b,i},
  \label{eq:recon1} \\
  \hat{h}^{(\ell_{m+1})}_{b,i}
  &= A_m \hat{h}^{(\ell_m)}_{b,i} + c_m
    + S_{m+1}^{-1}\hat{z}^{(m+1)}_{b,i}.
  \label{eq:reconm}
\end{align}
Equation~\eqref{eq:reconm} applies for $m = 1,\ldots,M-1$.
This chain means each later layer depends on the recursively reconstructed earlier layer,
and evaluation metrics are computed on the resulting $\hat{h}^{(\ell_m)}$ in original space.

\subsection{Online Multi-Layer Intervention}

{
During evaluation, we apply online hooks in layer order within a single forward pass. Let $\tilde{h}^{(\ell_m)}_{b,i}$ denote the activation that
arrives at selected layer $\ell_m$ during this modified forward pass, after any earlier hooks have
already changed the residual stream. Let $\hat{h}^{(\ell_m)}_{b,i}$ denote the activation written
back by the hook.

For a raw SAE, the hook at layer $\ell_m$ encodes the current activation
$\tilde{h}^{(\ell_m)}_{b,i}$ and writes the raw SAE reconstruction in original coordinates. This
is simple, but later raw SAEs may receive inputs produced by earlier replacements rather than by
the unmodified model, so their inputs can drift away from the clean activation distribution used
for training.

For a ReSAE, the hook uses the same affine decomposition as training, but conditions the prediction
term on the previous reconstructed selected layer. For $m>1$, the residual input to the layer
$\ell_m$ SAE is
\begin{equation}
  u_m
  = S_m\!\left(\tilde{h}^{(\ell_m)}_{b,i}
      - \bigl(A_{m-1}\hat{h}^{(\ell_{m-1})}_{b,i} + c_{m-1}\bigr)\right),
  \label{eq:online_input_resid}
\end{equation}
and the hook writes
\begin{equation}
  \hat{h}^{(\ell_m)}_{b,i}
  = A_{m-1}\hat{h}^{(\ell_{m-1})}_{b,i} + c_{m-1}
    + S_m^{-1}\hat{z}^{(m)}(u_m).
  \label{eq:online_resid}
\end{equation}
The anchor layer $\ell_1$ is reconstructed as in Equation~\eqref{eq:recon1}. Intuitively,
ReSAE hooks subtract the part of the current activation that is predictable from the previous
reconstructed selected layer before applying the SAE. Thus, if an upstream replacement mainly
changes a later activation through the learned affine relationship, that propagated change is
removed from the residual input. The later SAE then sees an input closer to its training target,
rather than a full activation shifted by earlier hooks.
}

\section{Experimental Setup}
\label{sec:setup}

\subsection{Evaluation Metrics}\label{sec:eval-metrics}

\paragraph{Cross-entropy and overinteraction.}
Let $\mathrm{CE}(S)$ be the language-model cross entropy after replacing the activations
at the layer set $S \subseteq L$ with SAE reconstructions during an online forward pass.
Let $\mathrm{CE}(\emptyset)$ denote the unmodified model. We define
\begin{equation}
  \begin{aligned}
    \Delta_S &= \mathrm{CE}(S) - \mathrm{CE}(\emptyset),\\
    \Delta_S^{\text{add}} &= \sum_{\ell \in S} \Delta_{\{\ell\}},\\
    \mathrm{OI}(S) &= \Delta_S - \Delta_S^{\text{add}}.
  \end{aligned}
  \label{eq:oi}
\end{equation}
{We introduce an additivity-residual metric, which we call overinteraction (OI), to quantify whether multi-layer replacement behaves like the sum of its single-layer effects. Specifically, we compare the observed CE degradation from jointly replacing a set of layers to the sum of CE degradations from replacing each layer independently. Values near zero indicate approximately additive composition, while deviations from zero indicate that the layers interact under joint replacement.}

\paragraph{Reconstruction quality.}
We report explained variance (EV) in reconstructed activation space, where after decoding each SAE
representation back into the model's original residual-stream coordinates, EV measures how
much variance in the clean activation is recovered. We also report cross-entropy degradation
under two replacement modes. In teacher-forced replacement, each SAE receives the clean
activation from an unmodified forward pass, isolating the quality of each reconstruction from
error propagation. In sequential-online replacement, hooks are applied in layer order during a
single modified forward pass, so later layers receive activations produced after earlier
replacements; this measures the intervention setting that matters when multiple SAEs are used
together.

\paragraph{Feature quality.}
We use feature-quality evaluations from SAE Bench~\cite{karvonen2025saebench}. Sparse probing
measures whether a small number of SAE latents can linearly recover task labels on
Bias-in-Bios, Amazon reviews, GitHub code, AG News, and Europarl. Targeted probe
perturbation (TPP) measures whether ablating latents associated with a target concept
selectively changes the corresponding classifier score, while spurious correlation removal
(SCR) tests whether ablating latents associated with a spurious attribute removes that
attribute without destroying task-relevant signal~\cite{karvonen2024evaluating}. We also
include first-letter splitting and absorption evaluations, following the observation that SAE
features can split or absorb related concepts across latents~\cite{chanin2025absorption}.
For SCR and TPP, we distinguish two intervention budgets. The \textit{global} budget ablates
the top $n$ latents across all selected-layer dictionaries, so the intervention uses $n$
features total. The \textit{per-layer joint} budget ablates the top $n$ latents separately in
each selected layer and applies all of those ablations in one forward pass, so for Pythia's
four selected layers it uses $4n$ total ablated features. 

\subsection{Models and Data}

We train SAEs on two transformer language models using the \texttt{monology/pile-uncopyrighted}
dataset. For \textbf{Pythia-1.4B-deduped}~\cite{biderman2023pythia}, we use the layer set
$\{0, 6, 12, 18\}$. For \textbf{Gemma-2-9B}~\cite{gemmateam2024gemma2}, a 42-layer model, we
use three layer sets for the layer-spacing study. We call the common distance between consecutive
selected layers the \textit{layer gap}: gap 4 denotes $\{2, 6, 10, 14, 18, 22, 26\}$, gap 6
denotes $\{2, 8, 14, 20, 26\}$, and gap 8 denotes $\{2, 10, 18, 26\}$.

All SAEs use the BatchTopK objective with dictionary size $N = 65{,}536$. Pythia sweeps
sparsity $k \in \{40, 80, 160, 320\}$, while Gemma uses $k = 160$ across the layer-set sweep.
Each SAE is trained for $5 \times 10^8$ tokens. The affine cross-layer regressions are
calibrated on $5 \times 10^5$ tokens before SAE training begins. We use Adam with learning
rate $3 \times 10^{-4}$, a 1000-step warmup, and 20\% learning-rate decay.

For every model and layer set, we compare two SAE families. The raw baseline trains standard
BatchTopK SAEs directly on the unresidualized activations. The residualized model trains SAEs
on pairwise affine residuals and reconstructs original-space activations with the regression
chain described above.

\section{Results}
\label{sec:results}

{
The main result is that ReSAEs recover cross entropy better under multi-layer replacement despite
having lower explained variance. We present the evidence in this order. First, we show that
residualization lowers raw reconstruction fidelity but reduces decoder redundancy. Second, we show
that this lower EV does not imply worse functional reconstruction: ReSAEs recover CE better in the
multi-layer setting that motivates the method. Third, we test whether the learned latents remain
useful for readout and targeted ablation. Finally, we repeat the central comparisons on
Gemma-2-9B to check whether the pattern persists at larger scale and across different layer
spacings.
}

\subsection{{Lower EV but Less Redundant Dictionaries}}
\label{sec:results:recon}

\begin{figure}[t]
  \centering
  \includegraphics[width=\linewidth]{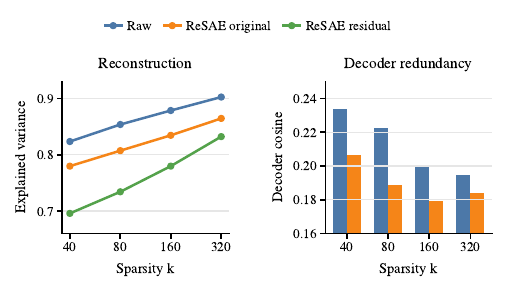}
  \caption{Pythia-1.4B reconstruction and decoder geometry. \textbf{Left}: raw original-space EV,
           ReSAE original-space EV, and ReSAE residual-target EV. \textbf{Right}: mean max decoder
           cosine similarity; lower means less redundant decoder directions.}
  \label{fig:reconstruction}
\end{figure}

{We first quantify the representation tradeoff introduced by residualization. Explained
variance (EV) measures how much clean activation variance is recovered, while decoder cosine
similarity measures whether different learned directions duplicate one another.}
Figure~\ref{fig:reconstruction} compares these quantities for the Pythia-1.4B sparsity sweep.
Raw SAEs have higher original-space EV at every sparsity level (0.823--0.902 vs.\
0.780--0.864), which is expected because ReSAEs train on the harder residual target and then
reconstruct through the affine chain. The residual-space EV of ReSAEs is lower still
(0.696--0.832), confirming that residualization removes a substantial linearly predictable
component before SAE training.

{However, ReSAEs are less redundant. They have lower mean max decoder cosine similarity at every
$k$, with the largest gap at $k=80$ (0.189 vs.\ 0.222). Thus EV favors the raw baseline, but
decoder geometry favors residualization: ReSAEs recover less raw activation variance, while
learning a less duplicated feature basis.}

\subsection{{Recovered CE Improves Despite Lower EV}}
\label{sec:results:ce_pythia}

\begin{figure}[t]
  \centering
  \includegraphics[width=\linewidth]{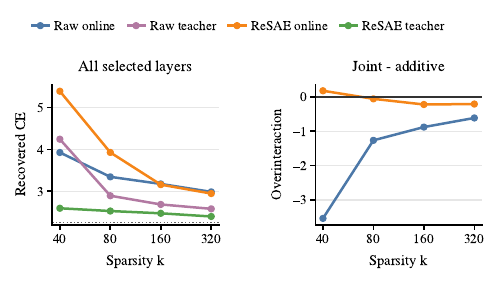}
  \caption{Pythia-1.4B recovered CE and overinteraction for all four selected layers.
           \textbf{Left}: recovered CE for raw and ReSAE replacement under both teacher-forced
           and sequential-online modes. \textbf{Right}: sequential-online
           overinteraction $\mathrm{OI}(L)$; values near zero indicate more additive composition.}
  \label{fig:pythia_ce}
\end{figure}

{We next evaluate whether lower EV translates into worse model behavior. Recovered cross entropy
measures the task-level effect of replacing model activations with SAE reconstructions, while
overinteraction measures whether multi-layer replacement behaves like the sum of the corresponding
single-layer replacements.}
Figure~\ref{fig:pythia_ce} shows recovered cross entropy when all four selected Pythia layers are
replaced. Teacher-forcing is shown for both raw SAEs and ReSAEs. The ReSAE teacher-forced curve is
best at every sparsity level, reaching CE 2.398 at $k=320$ compared with the clean-model CE of
2.260. Thus, despite lower original-space EV at every sparsity, ReSAEs recover the model's
cross-entropy behavior more faithfully when each layer is reconstructed from the clean activation.
Sequential online replacement is harder because later layers receive activations already modified
by earlier replacements. In this setting, raw SAEs are better at very low sparsity, but ReSAEs
overtake raw SAEs at $k=160$ and $k=320$.

{The overinteraction results help explain this gap between EV and functional replacement quality.
Raw SAEs have large negative overinteraction, especially at low $k$, meaning the joint intervention
is far from the additive prediction obtained from single-layer replacements. ReSAEs move much
closer to zero for $k\ge80$, indicating more predictable multi-layer composition. This suggests
that residualization is most useful once the sparse code has enough capacity to represent the
residual target: it may reconstruct less total variance, but its layerwise replacements compose
more reliably.}

\subsection{{Feature Readout and Targeted Ablation}}
\label{sec:results:pythia_saebench}

\begin{figure}[t]
  \centering
  \includegraphics[width=\linewidth]{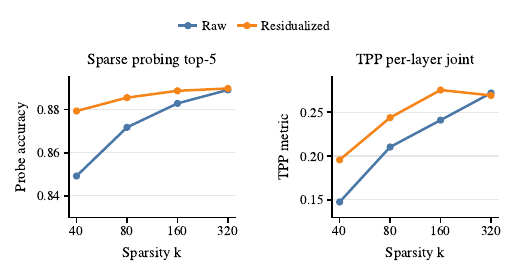}
  \caption{{Pythia-1.4B feature-level results.} \textbf{Left}: sparse probing accuracy using
           the top five SAE features across selected layers. \textbf{Right}: TPP under the
           per-layer joint budget. Higher is better in both panels.}
  \label{fig:pythia_saebench}
\end{figure}

{We then test whether residualized latents are useful beyond reconstruction. Sparse probing
measures linear readout from a small number of latents, TPP measures whether ablating
target-associated latents changes the intended probe score, and SCR measures whether spurious
information can be removed while preserving task-relevant signal.}
{Figure~\ref{fig:pythia_saebench} summarizes sparse probing and TPP on Pythia-1.4B. ReSAEs are
consistently better on sparse probing, with the largest gain at low sparsity ($0.879$ vs.\
$0.849$ top-5 accuracy at $k=40$) and near convergence by $k=320$. On TPP with a per-layer joint
budget, ReSAEs are also better through $k=160$ and essentially tied at $k=320$. These results
suggest that residualization improves feature usefulness for linear readout and targeted
perturbation, especially when the sparse budget is small.}

{Table~\ref{tab:scr_pythia} reports SCR. This stricter ablation test does not favor ReSAEs:
raw SAEs usually score higher, especially under the per-layer joint budget at larger $k$. This
indicates that residualized features are not uniformly better for every intervention objective;
in this setting, joint ablations of residualized features can also remove useful causal signal.}

\begin{table}[t]
  \centering
  \small
  \setlength{\tabcolsep}{4pt}
  
  \label{tab:scr_pythia}
  \resizebox{0.5\linewidth}{!}{%
  \begin{tabular}{lcccc}
    \toprule
    & \multicolumn{2}{c}{Global Budget} & \multicolumn{2}{c}{Per-Layer Joint} \\
    \cmidrule(lr){2-3}\cmidrule(lr){4-5}
    $k$ & Raw & ReSAE & Raw & ReSAE \\
    \midrule
    40  & $+0.035$ & $+0.032$ & $+0.028$ & $-0.012$ \\
    80  & $+0.045$ & $+0.002$ & $+0.007$ & $-0.040$ \\
    160 & $+0.035$ & $-0.022$ & $-0.013$ & $-0.124$ \\
    320 & $+0.011$ & $-0.007$ & $-0.043$ & $-0.077$ \\
    \bottomrule
  \end{tabular}
  }
  
  \caption{Pythia-1.4B SCR mean scores, averaged over SCR classes and ablation budgets $n$.
           Global ablates $n$ features total; per-layer joint ablates $n$ features per selected
           layer and applies the four layerwise ablations together.}
\end{table}

\subsection{{Larger-Scale and Layer-Spacing Checks}}
\label{sec:results:gemma_gap}

\begin{figure}[t]
  \centering
  \includegraphics[width=\linewidth]{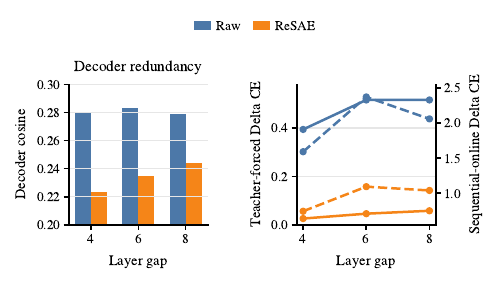}
  \caption{Gemma-2-9B layer-gap trends ($k=160$). \textbf{Left}: mean max decoder cosine
           similarity by layer gap; lower indicates less redundant decoder directions.
           \textbf{Right}: $\Delta$CE by layer gap. Solid lines use the teacher-forced left
           axis, while dashed lines use the sequential-online right axis.}
  \label{fig:gemma_ce}
\end{figure}

{We next test whether the same pattern holds in a larger model. The Gemma-2-9B sweep varies the
spacing between selected layers while keeping sparsity fixed, testing robustness to the choice of
multi-layer decomposition.}
Figure~\ref{fig:gemma_ce} summarizes the Gemma-2-9B layer-gap sweep. The left panel shows that
ReSAEs have lower decoder cosine similarity than raw SAEs at every gap, so residualization again
reduces decoder redundancy. The right panel shows cross-entropy degradation with separate axes for
teacher-forced and sequential-online replacement. ReSAEs are better in both modes at every gap:
teacher-forced $\Delta$CE stays near zero (0.026--0.058), and sequential-online $\Delta$CE is
roughly half of the raw baseline across the sweep (0.743--1.093 vs.\ 1.591--2.371). Thus the
central Pythia result---better recovered CE despite the reconstruction tradeoff---also appears in
Gemma.

\subsection{{Feature-Level Results at Larger Scale}}
\label{sec:results:probing}

{Finally, we repeat the feature-level checks on Gemma-2-9B.}
{For the gap-4 run, Table~\ref{tab:gemma_saebench} reports the results. ReSAEs improve sparse
probing at top-1, top-2, and top-5. TPP is mixed under the global budget but clearly better for
ReSAEs under the per-layer joint budget.}

\begin{table}[t]
  \centering
  \small
  \setlength{\tabcolsep}{5pt}
  \caption{{Gemma-2-9B feature-level results for the gap-4 layer set}
           $\{2,6,10,14,18,22,26\}$ at $k=160$.}
  \label{tab:gemma_saebench}
  \begin{tabular}{lcc}
    \toprule
    Metric & Raw & ReSAE \\
    \midrule
    Sparse probe top-1 & 0.786 & 0.840 \\
    Sparse probe top-2 & 0.817 & 0.859 \\
    Sparse probe top-5 & 0.865 & 0.901 \\
    TPP global budget & 0.121 & 0.118 \\
    TPP per-layer joint & 0.220 & 0.252 \\
    \bottomrule
  \end{tabular}
\end{table}
\section{Discussion}
\label{sec:discussion}

{
\subsection{Residualization trades raw EV for less redundant dictionaries}

The reconstruction results show that residualization should not be judged by original-space EV
alone. Raw SAEs reconstruct the full activation directly, so they naturally score higher on EV.
ReSAEs instead remove the part of the activation that is linearly predictable from an earlier
selected layer and train on what remains. This residual target has less easily recoverable
structure, which lowers EV, but it also reduces duplication among decoder directions. 

We note that ReSAEs reconstruct less of the raw activation but recover more of the model's cross entropy. EV measures variance recovery in a coordinate system the model itself does not privilege, whereas CE measures the functional aspect of the model. That these two quantities move in opposite directions is evidence that residualization steers capacity away from directions that carry forward predictable structure and toward directions that participate in the next layer's computation. This is the property a sparse dictionary should have for interpretability, where features correspond to what the model does and not to what it passively transmits.

\subsection{Residualization helps when the residual code is not too bottlenecked}

The intervention results show that the benefit of residualization depends on whether the sparse
code has enough capacity to represent the residual target. In Gemma, ReSAEs recover CE better than
raw SAEs across all tested layer gaps, in both teacher-forced and sequential-online replacement.
In Pythia, ReSAEs are strongest under teacher forcing and become competitive or better online once
$k$ is large enough, while low-$k$ online replacement still favors raw SAEs. This pattern suggests
a capacity cost, where if the residual code is too sparse, the affine chain cannot recover information
that the SAE failed to encode. Once that bottleneck is reduced, the residual target gives the
online intervention a cleaner object to replace and the layerwise replacements compose more
predictably.

\subsection{Feature-level gains are task-dependent}

The feature-level evaluations show that residualization improves some uses of latents but not all
of them. Sparse probing and TPP mostly favor ReSAEs, consistent with the view that removing
carried-forward structure concentrates capacity on information introduced at the selected layer.
SCR does not follow the same pattern on Pythia, where raw SAEs usually score higher, especially for
per-layer joint ablations. This suggests that residualized features can be more useful for readout
and targeted perturbation while still being less selective for the specific goal of removing a
spurious attribute without damaging the class signal. Thus, ReSAEs improve several
intervention-centered evaluations, but the best dictionary can still depend on the downstream
operation.

\subsection{Residualization is independent of the SAE objective}

The method should be viewed as a target construction step rather than as a commitment to
BatchTopK SAEs. Before training the SAE, we replace the raw activation target with the component
not explained by an affine prediction from an earlier selected layer. Once that target is defined,
the dictionary could be trained with other SAE objectives, including JumpReLU~\cite{rajamanoharan2024jumping},
p-Anneal~\cite{karvonen2024measuring}, or Matryoshka-style hierarchical training~\cite{bussmann2025matryoshka}.
Testing these combinations is an important direction for future work, since stronger SAE
objectives may recover more interpretable structure from the residual target.

\subsection{Affine residualization leaves open design choices}

The affine map is a useful first choice because it removes a simple, measurable form of
carried-forward structure. It is cheap to calibrate and keeps the residual target easy to
interpret, but it is unlikely to be the only useful choice. Low-rank or block-structured maps could
reduce memory cost, while nonlinear predictors might remove additional predictable structure at the
risk of discarding information useful for interpretation. For example, the attention module in a transformer linearly combines embeddings at each layer, where the main non-linearity enters through the softmax similarity function. Another
open question is layer selection, where the Gemma gap sweep suggests that distance between chosen layers
matters, but it does not yet tell us how to choose the best layer set for a task.
}

\section{Conclusion}
\label{sec:conclusion}

ReSAEs are a small change to the SAE training target with a large effect on multi-layer
behavior. Instead of having every layerwise dictionary relearn information already predictable
from an earlier selected layer, they train on the residual left by an affine cross-layer model and
reconstruct through that model at evaluation time. Across Pythia-1.4B and Gemma-2-9B, this reduces
decoder redundancy and improves several intervention-centered metrics, including recovered CE,
sparse probing, and targeted perturbation. The gains are not universal, and SCR in particular
shows that residualization can lose information useful for some causal-ablation objectives. Still,
the overall picture is encouraging since in multi-layer SAE interventions, treating the residual
stream as a coupled object is more faithful than treating each layer as an isolated training
problem.

\section*{Limitations}

The experiments cover two model families and a limited set of selected layers. Pythia is evaluated
with one four-layer set, while Gemma is evaluated across three regularly spaced layer sets at a
single sparsity level. This is enough to show that the effect is not tied to one architecture, but
not enough to claim that a particular layer gap or sparsity choice is generally optimal.

The residualization map is affine and is fit on a fixed calibration sample. This makes the method
simple and easy to analyze, but it also means any nonlinear predictable structure remains in the
SAE target. Conversely, a stronger predictor might remove information that is useful for
interpretability. The present experiments do not settle where that boundary should be.

The evaluation suite is intervention-heavy but still incomplete. Recovered CE, overinteraction,
sparse probing, TPP, and SCR each stress a different behavior, and they do not always agree. In
particular, Pythia SCR favors raw SAEs in several settings. This means ReSAEs should not be treated
as uniformly better dictionaries; they are better suited to some multi-layer replacement and
feature-use regimes than others.

Finally, the current implementation stores full affine maps between selected layers. That is
acceptable for the models studied here but could become expensive for wider models or denser layer
sets. Low-rank, block-structured, or otherwise compressed regression maps are likely needed before
this approach is convenient at much larger scale.

\section*{Acknowledgments}
This work was supported in part by the DARPA Young Faculty Award, the National Science Foundation (NSF) under Grants \#2127780, \#2319198, \#2321840, \#2312517, and \#2235472, \#2431561, the Semiconductor Research Corporation (SRC), the Office of Naval Research through the Young Investigator Program Award, and Grants \#N00014-21-1-2225 and \#N00014-22-1-2067, Army Research Office Grant \#W911NF2410360. Additionally, support was provided by the Air Force Office of Scientific Research under Award \#FA9550-22-1-0253, along with generous gifts from Xilinx and Cisco.

{
   \small
   \bibliographystyle{ieeenat_fullname}
   \bibliography{main}
}

\end{document}